\documentclass{article}

\usepackage{algorithm, algpseudocode}
\usepackage{amsfonts, amsmath, amsopn, amsthm, epsfig}
  \numberwithin{equation}{section}
\usepackage{graphicx, epstopdf}
\usepackage{hyperref}
\usepackage{subfigure}
\usepackage{appendix}
\usepackage{authblk}

\usepackage{calc}
\usepackage[top=2.5cm,bottom=2.3cm,inner=2.3cm,outer=2.3cm,bindingoffset=0.5cm,footskip=0.55cm]{geometry}
\theoremstyle{remark}

\newenvironment{lemma*}[2][Lemma]{\par\bgroup{\bfseries #1\ #2. }\it\ignorespaces}{\egroup}

\title{HiCOMEX: Facial Action Unit Recognition Based on Hierarchy Intensity Distribution and COMEX Relation Learning}

\author[1]{Ziqiang Shi\thanks{Corresponding author: shiziqiang@cn.fujitsu.com}}
\author[1]{Liu Liu}
\author[1]{Zhongling Liu}
\author[1]{Rujie Liu}
\author[2]{Xiaoyu Mi}
\author[2]{Kentaro Murase}

\affil[1]{Fujitsu Research and Development Center, Beijing, China}
\affil[2]{Fujitsu Laboratories Ltd., Kawasaki, Japan}

\newcommand{\BALD}{\begin{aligned}}
\newcommand{\EALD}{\end{aligned}}
\newcommand{\BALDS}{\begin{aligned*}}
\newcommand{\EALDS}{\end{aligned*}}
\newcommand{\BCAS}{\begin{cases}}
\newcommand{\ECAS}{\end{cases}}
\newcommand{\BEAS}{\begin{eqnarray*}}
\newcommand{\EEAS}{\end{eqnarray*}}
\newcommand{\BEQ}{\begin{equation}}
\newcommand{\EEQ}{\end{equation}}
\newcommand{\BIT}{\begin{itemize}}
\newcommand{\EIT}{\end{itemize}}
\newcommand{\BMAT}{\begin{bmatrix}}
\newcommand{\EMAT}{\end{bmatrix}}
\newcommand{\BNUM}{\begin{enumerate}}
\newcommand{\ENUM}{\end{enumerate}}

\newcommand{\BA}{\begin{array}}
\newcommand{\EA}{\end{array}}

\date{}

\begin{document}

\maketitle

\renewcommand{\thefootnote}{\fnsymbol{footnote}}


\begin{abstract}
  The detection of facial action units (AUs) has been studied as it has the competition due to the wide-ranging applications thereof.
  In this paper, we propose a novel framework for the AU detection from a single input image by 
  grasping the \textbf{c}o-\textbf{o}ccurrence and \textbf{m}utual \textbf{ex}clusion (COMEX) as well as 
  the intensity distribution among AUs. Our algorithm uses facial landmarks to detect the features of local AUs.
  The features are input to a bidirectional long short-term memory (BiLSTM) layer for learning the 
  intensity distribution. Afterwards, the new AU feature  continuously passed through
a self-attention encoding layer and a continuous-state modern Hopfield layer for learning the COMEX relationships.
  Our experiments on the challenging BP4D and DISFA benchmarks without any external data or pre-trained models yield F1-scores of 63.7\% and 61.8\% respectively, 
  which shows our proposed networks can lead to performance improvement in the AU detection task. 
\end{abstract}

\section{INTRODUCTION}

Faces carry a large amount of information about our psychology and emotional states at all time, thus we can quantify mental status based on 
facial analysis.
An action unit, defined by the Facial Action Coding System (FACS)~\cite{ekman1997face}, represents the basic facial muscle 
movement or expression change, 
and the aim of facial action unit detection is to detect and estimate the
occurrence of certain individual facial muscle movements by determining whether the AUs appear or not. 
AU detection has a vast range of applications, for example, the improvement of labour productivity by 
taking care of employee’s 
psychology and 
improving their motivation, or estimating customer’s satisfaction and improving their purchasing 
motivation (digital marketing).
In these cases, a detail description of AU occurrence needs to be estimated from the facial images to complete the subsequent tasks,
e.g., facial micro-expression analysis, thus it is a problem that must be solved  achieve satisfactory performance in 
subtle facial signal analysis tasks.

Many techniques based on deep learning have been proposed for this task. 
Zhao et al.~\cite{zhao2016deep} 
presented deep regional feature learning and 
multi-label learning modules in a unified architecture for facial AU detection. 
Li et al.~\cite{li2017eac} proposed the use of an EAC-Net that can learn both feature enhancing and region cropping functions simultaneously.
Corneanu et al.~\cite{corneanu2018deep} proposed a deep structured inference network (DSIN) for AU detection
in two stages: in the first stage, learned local and global features are combined for initial estimations of AU occurrences; 
in the second stage, structural inference is used to capture AU relationships by passing information among the initial estimations  
for final predictions of AU occurrences.
Niu et al.~\cite{niu2019local} improved the AU recognition by using facial shape information extracted from landmarks, although
they did not extract a feature map for each AU or utilize any AU relationships.
Shao et al.~\cite{shao2019facial,shao2020jaa} developed an end-to-end framework  for joint facial AU detection and face alignment, 
which can 
contribute to each other by sharing features and initializing the attention maps with the results of face alignment.

\begin{figure*}[th]
  \centering
  \subfigure[Co-occurrence of AU1, AU2, and AU5.]{
  \begin{minipage}[t]{0.25\linewidth}
  \centering
  \includegraphics[width=0.9in]{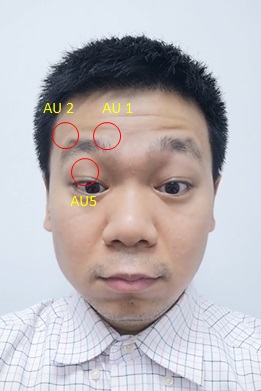}
  \label{au_1_2_5}
  \end{minipage}%
  }%
  \subfigure[Co-occurrence of AU4, AU7, and AU9.]{
  \begin{minipage}[t]{0.25\linewidth}
  \centering
  \includegraphics[width=0.9in]{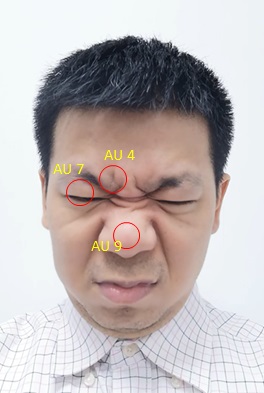}
  \label{au_4_7_9}
  \end{minipage}%
  }%
  \subfigure[Sample of AU12.]{
    \begin{minipage}[t]{0.25\linewidth}
    \centering
    \includegraphics[width=0.9in]{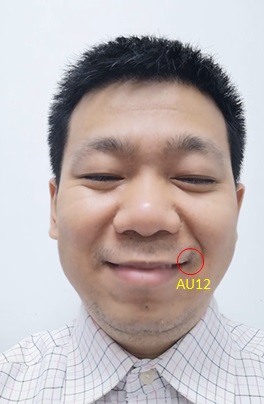}
    \label{au_12}
    \end{minipage}%
    }%
    \subfigure[Sample of AU15.]{
      \begin{minipage}[t]{0.25\linewidth}
      \centering
      \includegraphics[width=0.9in]{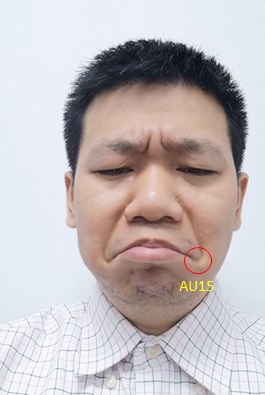}
      \label{au_15}
      \end{minipage}%
      }%
  \centering
  \caption{Examples of co-occurrence and mutually exclusive relations between AUs. A surprised expression may cause AU1, AU2, and AU5 to appear together. 
 An angry expression may cause AU4, AU7, and AU9 to appear together. Generally AU12 and AU15 will not appear at the same time.}
 \label{co_occurrence_example}
  \end{figure*}

Although these methods are effective, there are still two difficulties in facial AU detection that have 
not been resolved. 
Firstly, the subtle  change of local facial appearance must be accurately measured to judge whether an AU happens or not: this remains 
challenging. Individuals may have different levels of expressiveness due 
to their physical characteristics, which further increases the difficulty in detecting subtle facial changes.
The second shortcoming of previous methods is that they do not fully explore
 \textbf{c}o-\textbf{o}ccurrence and \textbf{m}utual \textbf{ex}clusion (COMEX) of different AUs for overall AU prediction. 
 AUs are not mutually independent of each other, since facial expressions are closely related to AU occurrences~\cite{zhao2016joint,wang2013capturing}.
 As with the examples shown in Fig.~\ref{co_occurrence_example}, some AUs usually happen 
 simultaneously, while some AUs never appear at the same time, which means several AUs can usually be active at the 
same time and certain AU combinations are more likely to happen than others.

   \begin{figure*}[th]
      \centering
      \hspace{-5mm}
      \includegraphics[width=1.0\linewidth]{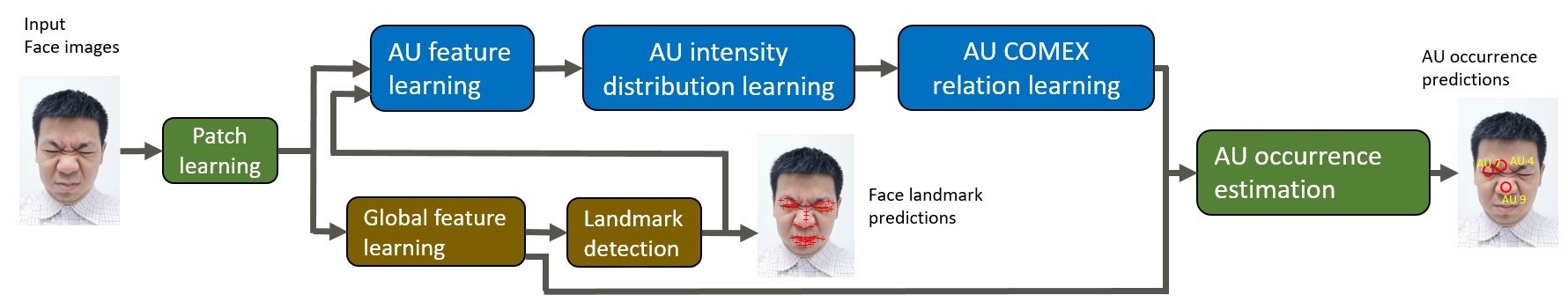}
      \hspace{-5mm}
    \caption{
      Framework of the proposed method. First, the patch learning module is used to convert
       the input facial signal into its patch learning feature. Then, the global facial feature
       and local AU features
        are extracted based on this patch learning feature.  
        Before concatenating these two features, 
        the AU features are recombined through the interaction 
        among AUs to maximise the characterisation efficacy. The AU occurrences
         are predicted based on the concatenated representations.
        }
      \label{detection_pipeline_simple}
      \end{figure*}
  
  To overcome the above limitations, we propose an end-to-end deep neural network approach called 
  hierarchical local AU region  intensity distribution and COMEX 
  relation learning (HiCOMEX) for AU detection as shown in Fig.~\ref{detection_pipeline_simple}. 
  HiCOMEX can dynamically 
  calculate the intensity distribution and COMEX relationships among all AUs based on each input image, 
  and strengthen the expressive ability of AU feature maps.

  The key contributions of this paper are demonstrated as follows:
  \begin{itemize}
    \item HiCOMEX is the first study to adapt single image to BiLSTM to reflect intensity 
    distribution among AUs into AU feature maps.
    \item HiCOMEX is the first study to employ the self-attention mechanism  and the modern Hopfield network~\cite{ramsauer2020hopfield} to 
    address the AU COMEX relationship modelling for AU detection. 
    It can use this relationship to enhance the representation ability of AU features. 
     In other words, these modules naturally make HiCOMEX  independent of a person's 
     identity. 
    \item HiCOMEX can achieve state-of-the-art performance on the notable BP4D and DISFA benchmarks.
    \end{itemize}
  
  The remainder of this paper is organised as follows:  Section~\ref{sec:related_work} introduces related work.
  Section~\ref{sec:lrrl} describes our proposed 
  HiCOMEX and 
  AU detection algorithm. 
  The experimental set-up and results are presented in Section~\ref{sec:experiments}. We conclude this paper in 
  Section~\ref{sec:conclusion}.

\section{Related work}
\label{sec:related_work}

Our HiCOMEX is closely related to existing deep learning-aided facial AU detection methods as well as AU modelling
with relation (graph) learning methods, since we combine
both AU detection models and relationship (graph) learning models.

Chu et al.~\cite{chu2017learning} proposed a hybrid deep learning framework that used the strengths of CNNs and LSTMs to
model and utilise both spatial and temporal cues.
Ertugrul et al.~\cite{ertugrul2019pattnet} proposed a simple sigmoidal attention mechanism for weighting the local facial patches to detect specific AUs. 
Li et al.~\cite{li2019semantic} combined the fixed knowledge graph of AU correlation into a convolution neural network to enhance AU detection.
Reale et al.~\cite{reale2019facial} extracted local AU features in the 3-d space based on cloud input.
Tu et al.~\cite{tu2019idennet} implemented an identity-aware architecture of multi-task network cascades where one task is to extract identity information, the 
other is to subtract that identity information and undertake AU detection. 
Fan et al.~\cite{fan2020facial} presented a new learning framework that automatically
learns the latent relationships between AUs via establishing semantic
correspondences between feature maps.

Our proposed HiCOMEX is a more flexible solution that can treat all AU feature maps as a sequence data or a graph data at the 
same time. We can dynamically calculate the relationship between AUs based on the input image in real-time, and use this relationship 
to improve the characterisation ability of AU feature maps. In other words, our method is naturally independent of a person's identity.

  

  \section{AU recognition with HiCOMEX}
  \label{sec:lrrl}

  
  \subsection{AU Feature Representation Learning}
  \label{sec:aufeature}
  
\begin{figure}[!h]
    \centering
    \hspace{-5mm}
    \includegraphics[width=1.0\linewidth]{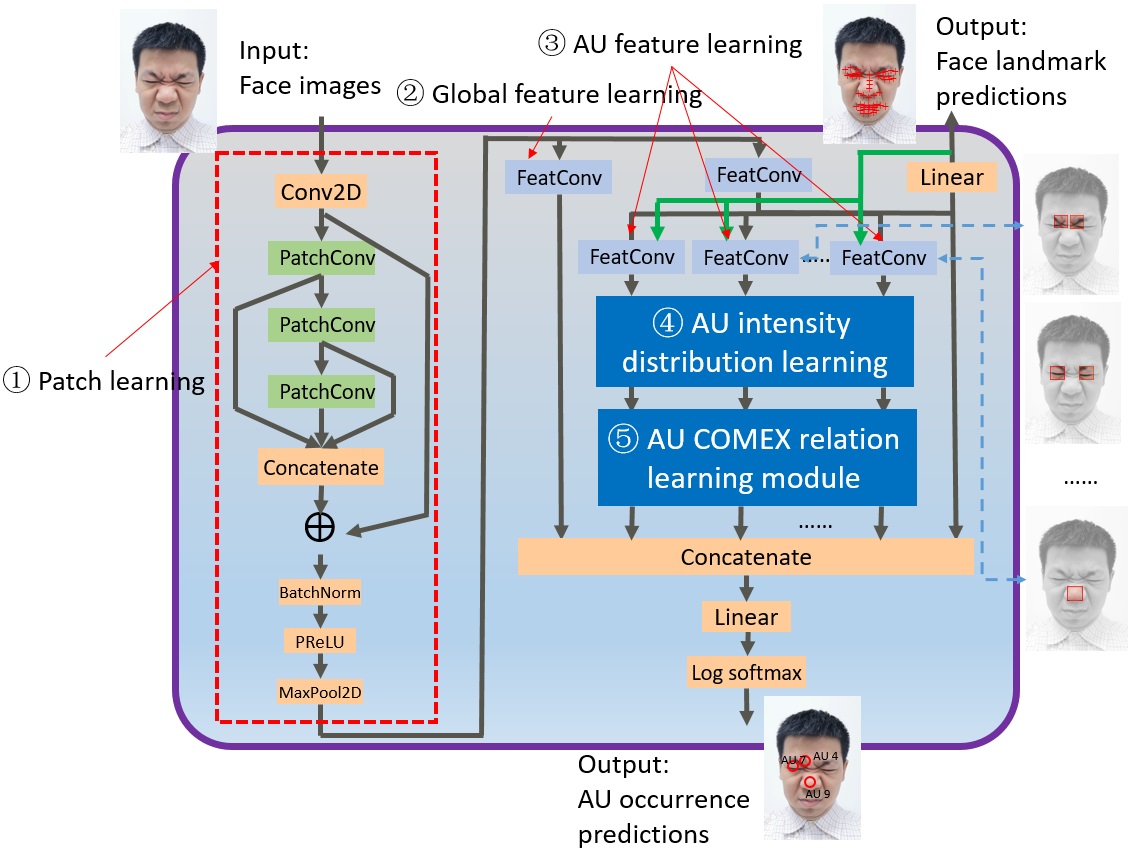}
    \hspace{-5mm}
    \caption{
  The architecture of HiCOMEX consists of 1) patch learning, 
    2) global facial feature learning, 3) 
    AU feature learning, 
    4) AU  intensity distribution learning (see Fig.~\ref{rl_w_bilstm}), 5) 
    AU COMEX relation learning (see Fig\ref{rl_w_hopfield})
    6) AU occurrence prediction, and 7) facial landmark estimation. 
    }
    \label{detection_pipeline}
    \end{figure}

  Figure~\ref{detection_pipeline} illustrates the detailed framework of our proposed approach to achieve this goal.
  Firstly the patch learning module~\cite{zhao2016deep} maps a facial image to its latent feature representation 
  called 
  \emph{patch features}
  with precise features 
  of local regions. 
  To improve localisation and classification, `PatchConv'~\cite{zhao2016deep}  is adopted 
  here for the patch learning. 
  With the PatchConv structure, the initial feature map obtained by 2-d original image is divided spatially into several parts, which
  are then input separately to the following branches where independent but identical convolution operations are implemented.
   After that, the feature maps 
  from these branches are reconcatenated to form the final feature map. 
For details of `PatchConv' and `FeatConv', please refer to the work of ~\cite{zhao2016deep}.

  Then inspired by DSIN~\cite{corneanu2018deep}, 
  the \emph{global facial feature}, local \emph{AU features}, and facial \emph{landmark features}
    are extracted based on patch features.
  All global facial feature, facial landmark feature, 
  and local AU feature 
  representations are extracted 
  through several layers of `FeatConv', which consists of 2 Conv2Ds, 
  BatchNorms, PReLUs, and a 2-D max-pooling (Max Pool2D). 
  
  \subsection{Local AU Feature Extraction}
  \label{sec:localaufeature}
  
\begin{figure}[th]
      \centering
      \hspace{-5mm}
      \includegraphics[width=1.0\linewidth]{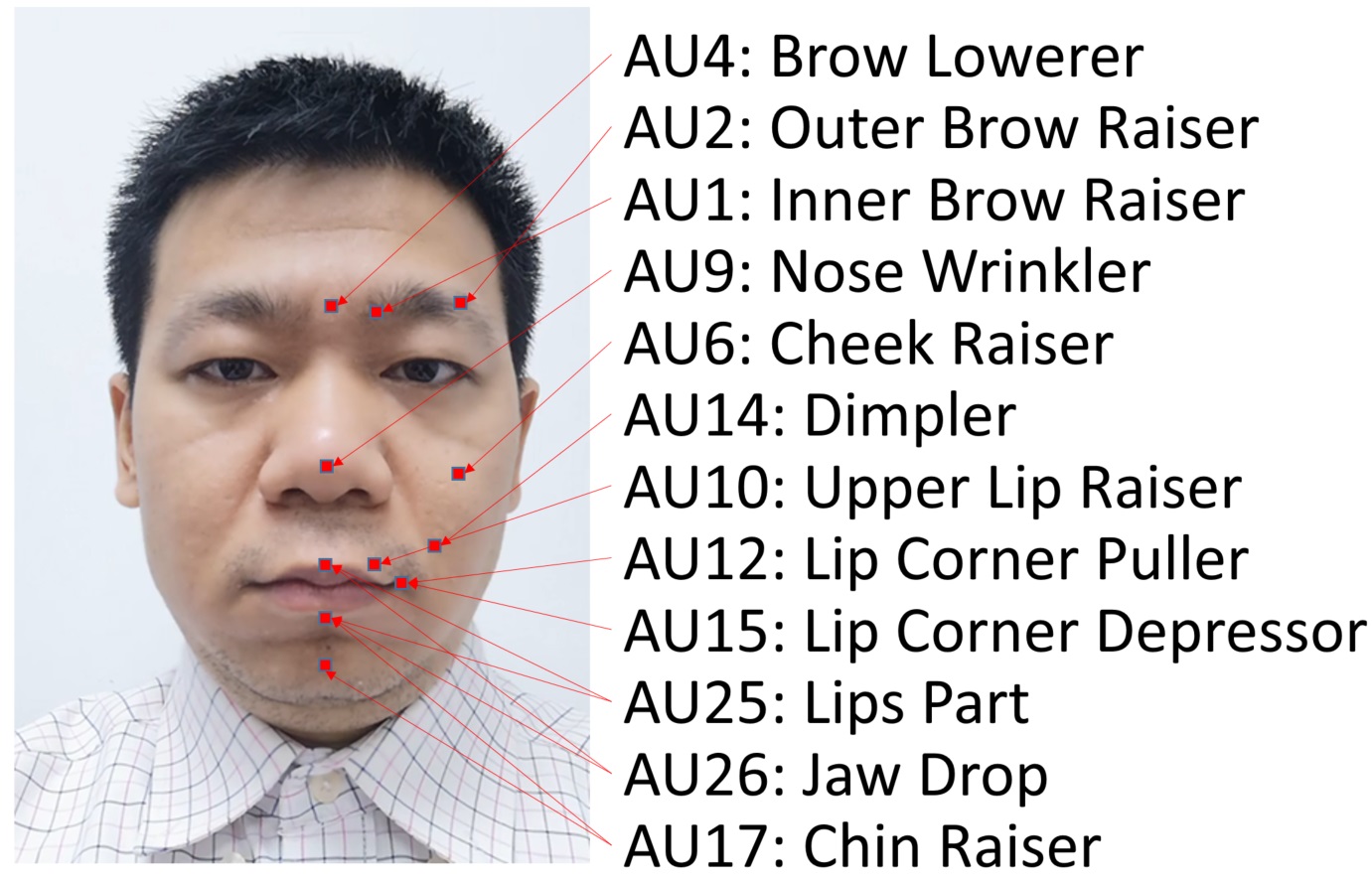}
      \hspace{-5mm}
      \caption{
      The centers of AUs.
      }
      \label{au_centers}
      \end{figure}

  In HiCOMEX, local AU features play a more important role than the global counterpart, 
  due to their role in later  intensity distribution and COMEX relationship learning 
  and their contribution to final feature representation. In our implementation, the local AU features 
  are not obtained by an extra 
  process, instead, they are cropped and refined from the certain global feature map according to 
  the local facial area of 
  each AU
  , as shown 
  in Fig.~\ref{au_centers}.
  As we know, the facial region which causes the occurrence of each AU is manually defined, e.g., AU1 and AU2 come respectively 
  from muscle movement in the inner brow and outer brow area. With this prior knowledge, we may further define the approximate region for 
  each AU, as illustrated in Fig.~\ref{au_centers}.   
The landmark feature will be used to predict the
locations of the landmark points.
According to the locations of the landmark points, we can roughly determine the facial area 
corresponding to each AU and will then extract the corresponding AU feature map 
according to the local area of the face corresponding to the AU.
To mine and use the relationship among AUs, all local AU features will pass through an AU 
intensity distribution learning module and a COMEX relation module  to enhance their representation ability, 
and the output contains the refined new 
local AU features, which will be concatenated with the global facial feature and landmark feature map, 
for final AU recognition.

  \begin{figure}[!h]
    \centering
    \hspace{-5mm}
    \includegraphics[width=1.0\linewidth]{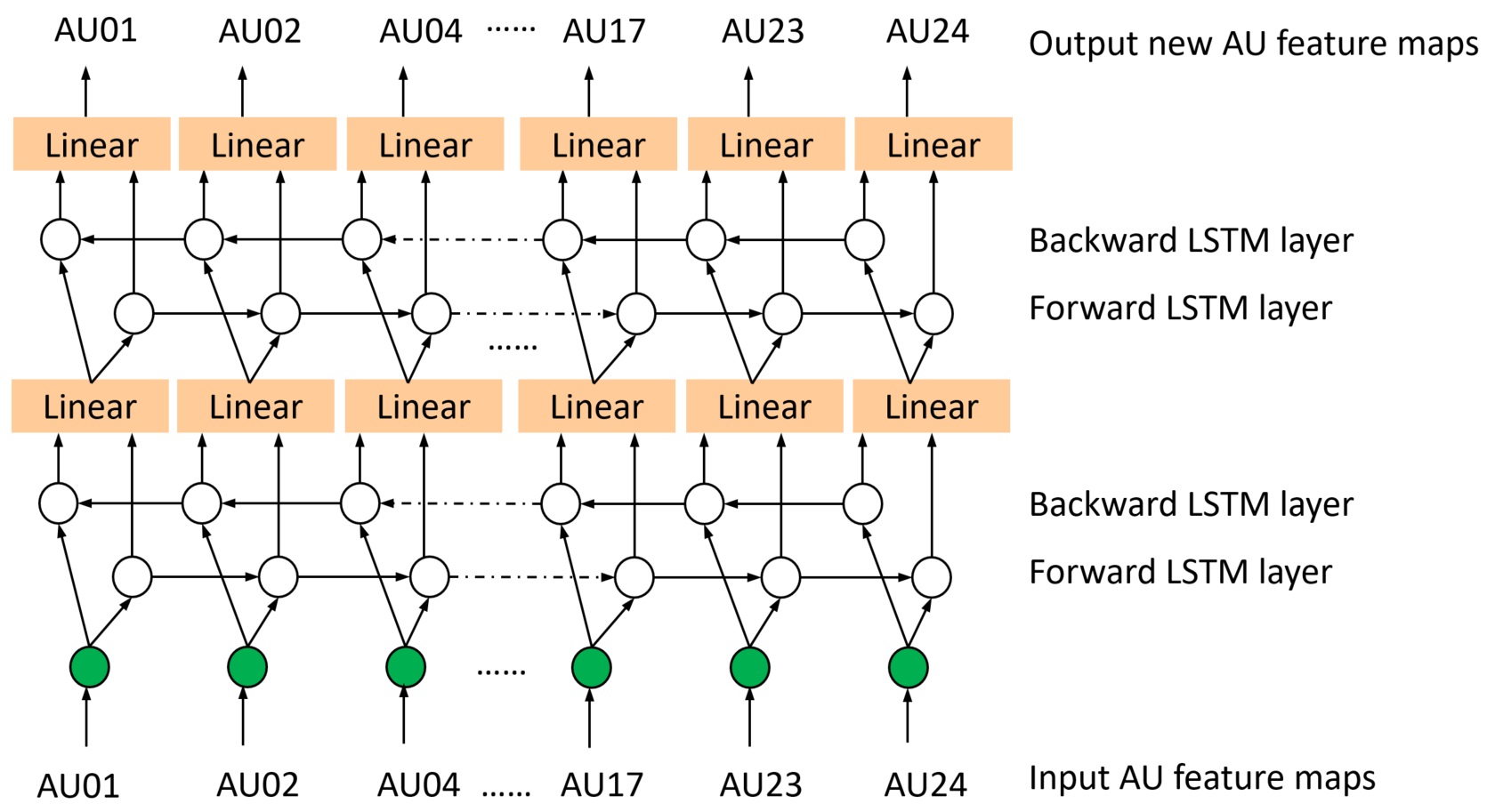}
    \hspace{-5mm}
      \caption{
      AU intensity distribution  learning with BiLSTM. The feature maps corresponding to all AUs are input 
        into BiLSTM as a sequence, and the output is the AU feature maps after recombination and 
        transformation. The reason for using BiLSTM in two directions is to help all AUs before 
        and after can interact and influence  with each other.
    }
    \label{rl_w_bilstm}
    \end{figure}

  \subsection{AU Intensity Distribution Learning}
  \label{sec:BiLSTM}
  
  In this study we adapt single image into BiLSTM to handle intensity distributions resulting from  intensity differences among AUs due 
  to the asynchronicity of occurrence of AUs. When a person is adopting a certain expression, due to the physical characteristics of the facial muscles, some AUs usually 
  appear asynchronously. Within a labelled AU data set consisting of many single images, AUs will follow a  certain statistical intensity distribution in each AU co-occurrence 
  and mutual exclusion pattern, therefore, the AU feature maps from a single image must include information of the intensity difference that represents an intensity 
  distribution pattern. We propose to learn this intensity distribution using BiLSTM.
  
  As shown in Fig.~\ref{rl_w_bilstm}, we input AU feature maps 
  obtained from a single image by the previous AU feature extraction into 
  BiLSTM as sequence data. Any predefined AU order, such as natural order 
  AU1, AU2, ..., and AU24, can be input into BiLSTM, since knowledge about the order of AUs is unknown or undecided. Bi-directional LSTMs are used to allow 
  information to propagate in both directions  to interact the AU feature maps in the sequence back and forth. Intensity distribution learning is  data-driven:
   in one image, the intensities of AUs are different because the AUs occur in different orders. The intensity differences due to the asynchronous AU 
  occurrence can be learned through BiLSTM, while at the same time this method will enhance high-intensity AUs and weaken low-intensity AUs from the viewpoint of classifying 
  the intensity distribution patters. In this study, we adapted each single image to BiLSTM for the first time to reflect the intensity distribution of AUs into AU feature maps. BiLSTM 
  can create the new AU features with intensity distributions to enhance their discriminative ability with regard to AUs.

\begin{figure}[th]
  \centering
  \hspace{-5mm}
  \includegraphics[width=1.0\linewidth]{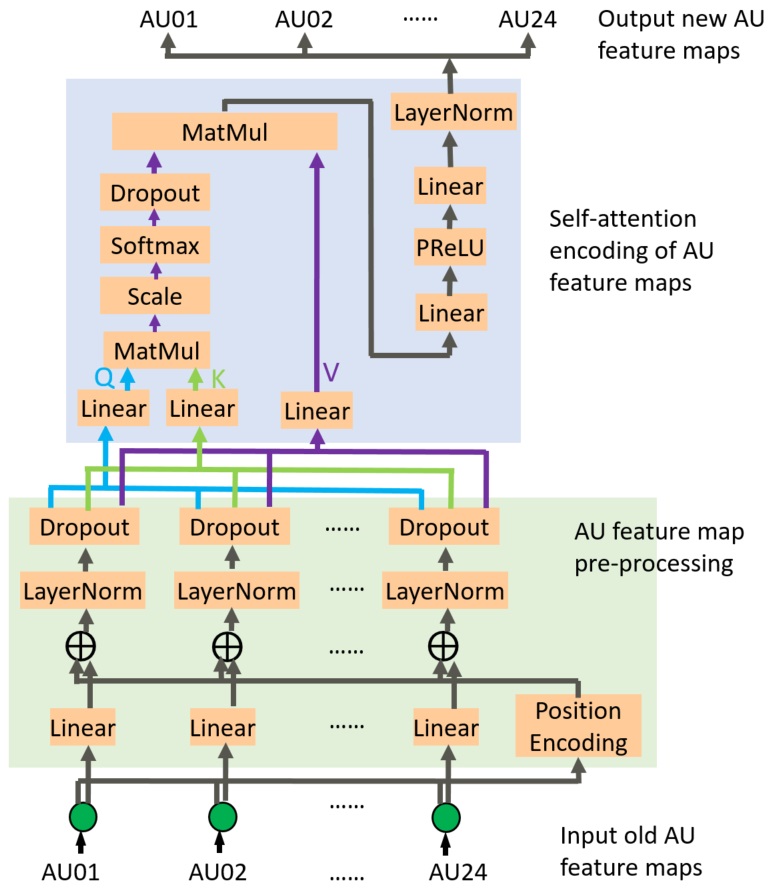}
  \hspace{-5mm}
  \caption{
    AU COMEX relation learning with self-attention encoding. The encoding is composed of two parts. The first is to 
    perform the pre-processing of a linear transformation on the old AU feature maps, and in this process the coded 
    position information of the AU feature maps are added. The second is the self-attention encoding layers, which can be 
    described as mapping a query and a set of key-value pairs to an output, where the query, key, value, and output are all vectors. 
    The output is calculated as a weighted sum of values, where the weight assigned to each value is derived by using the corresponding 
    key query compatibility function.
  }
  \label{rl_w_self_attention}
  \end{figure}

  \subsection{AU COMEX Relation Learning with Self-Attention Encoding}
  \label{sec:Hopfield}

    As described in Section~\ref{sec:BiLSTM}, the intensity distribution information among AU features is explored by BiLSTM as
   a sequential structure, 
  however, the COMEX relationship among AUs, which may contribute more to 
  enhancing the discriminative ability of AU features  is not involved at all. To alleviate this weakness, a 
 graph structure is naturally adopted to describe  such 
 a relationship between AUs~\cite{fan2020facial,wang2013capturing,li2019semantic}.

  In the graph structure, each AU is represented as a graph node, while the relationship such as the 
  probability of co-occurrence among 
  two AUs is represented as the graph edge. The higher the probability that two AUs appear together, 
  the larger the corresponding 
  edge value, and vice versa. For example, the value of the edge that connects AU12 and AU15 should be a 
  very small value, because these 
  two AUs are conflicting and never appear at the same time. 
   In this section we use  self-attention~\cite{vaswani2017attention} 
to model both the graph and sequence structure of AUs simultaneously.

As shown in Fig.~\ref{rl_w_self_attention}, the AU feature map first passes through the pre-processing module, 
which contains fully connected linear units (Linear), layer normalisation  (LayerNorm)~\cite{ba2016layer}, 
dropout (Dropout)~\cite{srivastava2014dropout}, and a position encoding module. The position encoding is undertaken as follows
\begin{eqnarray}
  pe(pos,2i)&=& \sin(pos/10000^{2i/d}) \\
  pe(pos,2i+1)&=& \cos(pos/10000^{2i/d})
\end{eqnarray}
where $pos$ is the position of the AU feature in the input AU graph and 
$i$ denotes the dimension in the linearly-transformed AU features, and $d$ is the output 
dimension of the linear component of the pre-processing  module. This $pe$ vector is added to the
linearly-transformed AU features. After the preprocessing module, we acquire  input AU 
feature maps for the  self-attention encoding module.

Several identical self-attention encoding layers can be used in our framework. In each self-attention encoding layer, 
the input consists of queries and keys of dimension $d_k$, and values of dimension $d_v$. Here queries, keys, and values are different 
avatars of the same AU feature map, although  modified  through different linear transformations.
Then we compute the attention function on a set of queries simultaneously, packed together
into a matrix $Q$. The keys and values are also packed together into matrices $K$ and $V$. We compute
the matrix of outputs as:
\begin{equation}
  \text{Self-Attention}(Q,K,V)=\text{softmax}(\frac{QK^T}{\sqrt{d_k}})V.
  \label{sa_update}
\end{equation}
Indeed $\frac{QK^T}{\sqrt{d_k}}$ describes the distances  between all AU node pairs in the graph. Thus $\text{Self-Attention}(Q,K,V)$ is indeed
the weighted sum of AU features according to the relations computed by $\frac{QK^T}{\sqrt{d_k}}$, thus result in new AU representations. This means
that co-occurring  AUs will enhance each other and mutually exclusive AUs will reduce the probability of their co-occurrence.

  \subsection{AU COMEX Relation Learning with a Continuous Hopfield Layer}
  \label{sec:Hopfield}

  \begin{figure}[th]
    \centering
    \hspace{-5mm}
    \includegraphics[width=1.0\linewidth]{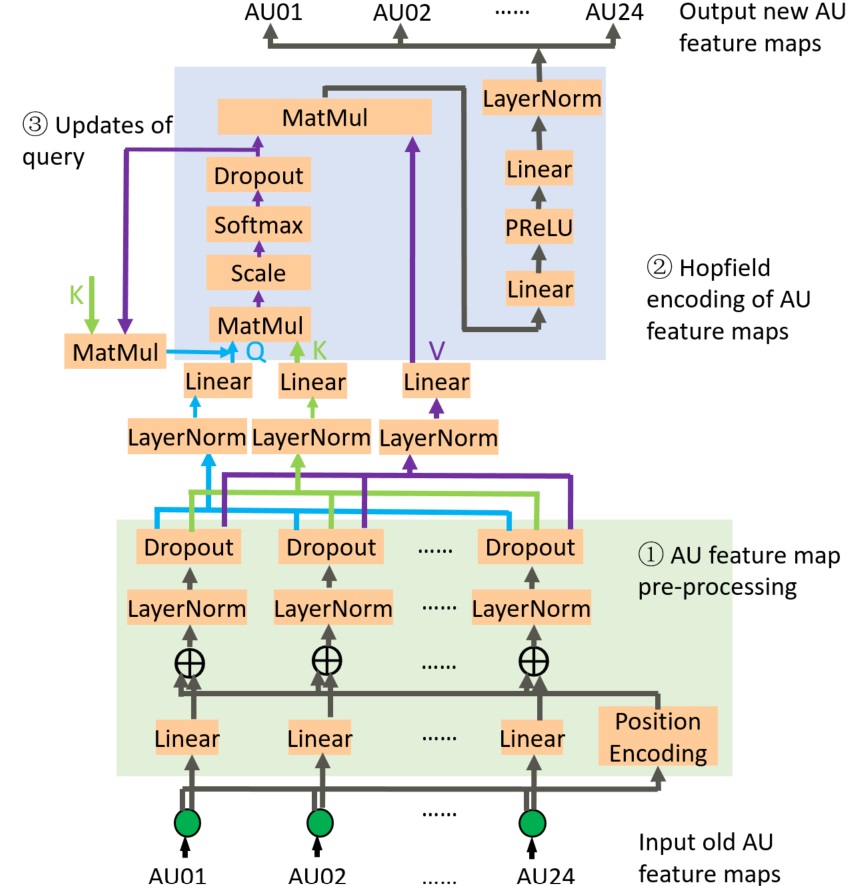}
    \hspace{-5mm}
    \caption{
    AU COMEX relationship learning with a Hopfield layer. The Hopfield relationship learning consists of three parts: 1)pre-processing of a linear 
 transformation on the old AU feature maps, (in this process the coded 
  position information of the AU feature maps is added); 2) self-attention; 3) updates of the query.
    }
    \label{rl_w_hopfield}
    \end{figure}
  
Recently Ramsauer et al.~\cite{ramsauer2020hopfield} proposed the use of a modern Hopfield network with continuous states. 
  This novel modern Hopfield network~\cite{ramsauer2020hopfield} generalises  the 
  original Hopfield network with continuous states, 
  which provides a good choice for relationship learning due to its intrinsic graph structure.  
  A simple and effective training method is proposed, where the basic principle of the training method is to obtain the attention 
  by constantly providing queries to the network, akin to  the self-attention mechanism.
Through this discovery, they have improved the existing 
self-attention mechanism. Here we also try to use this new model to learn the COMEX relationship between AUs and strengthen the characterisation ability of AU feature maps.

As shown in Fig.~\ref{rl_w_hopfield}, the difference in modern Hopfield network from a self-attention system is that 
the updating  formula in Eq.~(\ref{sa_update}) can be applied iteratively 
to the initial state of each Hopfield layer header.  After the last update, the new state will be projected to the resulting pattern. 
  Therefore, the Hopfield layer allows multiple update steps in a forward direction without changing the number of parameters. 
  The number of update steps per Hopfield header can be specified.
  In addition, a threshold can be set for the number of updates based on the size of each Hopfield header.

  \subsection{Losses}
  \label{sec:loss}

  
    HiCOMEX has two kinds of outputs (as shown in Fig.\ref{detection_pipeline}), 
    one is AU occurrence prediction, which is 
    a 0/1 prediction for each AU, the other is landmark prediction, which are the coordinates of each landmark. 
  In the present paper, we use two kinds of losses, namely the cross-entropy loss of AU 
  occurrence based on the concatenated global and local feature maps and the Euclidean loss of 
  landmark prediction accuracy based on the landmark feature maps.

\subsection{Two-stage Training}
\label{sec:training}

During the experiment, we found that the two-step training method works better. That is, firstly we train the pipeline in Fig.~\ref{detection_pipeline} without 
the local AU relationship-learning module to acquire the best patch learning, global facial representation learning, and local AU representation learning modules. 
Then the local AU relation learning module is again  added to result in the HiCOMEX, and the parameters the best
 patch learning, global facial representation learning, and local AU representation learning modules are fixed, 
only training local AU relationship learning, AU occurrence prediction, and facial landmark estimation modules.
The training algorithm we use is SGD, with a momentum of 0.9 and a weight decay 
of 0.0005. We train HiCOMEX with an initial learning rate of 0.01 for up to 30 epochs, where the learning rate is multiplied by a 
coefficient of 0.3 for every two epochs.

  
  \begin{table*}[th]
    \caption[f1_bp4d]{F1-score(\%)  in a comparative study of different state-of-the-art detection methods on the 
    BP4D dataset.}\label{tab:f1_bp4d}
    \centering
    \begin{tabular}{c|c|c|c|c|c|c|c|c|c|c|c|c|c}
    \hline
    \hline
    AU & 1 & 	2	 & 4	 & 6	& 7	& 10	& 12	& 14& 	15	& 17	& 23	& 24& Avg\\
    \hline
    \hline
    LSVM &23.2 &	22.8	& 23.1&27.2&47.1&77.2&63.7&64.3&18.4&33.0&19.4&20.7&35.3\\
    JPML &32.6&25.6&37.4&42.3&50.5&72.2&74.1&65.7&38.1&40.0&30.4&42.3&45.9\\
    DRML &36.4&41.8&43.0&55.0&67.0&66.3&65.8&54.1&33.2&48.0&31.7&30.0&48.3\\
    CPM &43.4&40.7&43.3&59.2&61.3&62.1&68.5&52.5&36.7&54.3&39.5&37.8&50.0\\
    EAC-Net &39.0&35.2&48.6&76.1&72.9&81.9&86.2&58.8&37.5&59.1&35.9&35.8&55.9\\
    DSIN&51.7&40.4&56.0&76.1&73.5&79.9&85.4&62.7&37.3&62.9&38.8&41.6&58.9\\
    CMS&49.1&44.1&50.3&79.2&74.7&80.9&88.3&63.9&44.4&60.3&41.4&51.2&60.6\\
    LP-Net&43.4&38.0&54.2&77.1&76.7&83.8&87.2&63.3&45.3&60.5&48.1&54.2&61.0\\
    ARL &45.8&39.8&55.1&75.7&\textbf{77.2}&82.3&86.6&58.8&47.6&62.1&\textbf{47.4}&\textbf{55.4}&61.1\\
    SCC &48.2&50.3&55.8&79.9&73.6&\textbf{88.1}&89.9&\textbf{68.7}&42.2&\textbf{65.2}&41.6&42.0&62.1\\
    J$\hat{\text{A}}$A-Net &\textbf{53.8}&47.8&58.2&78.5&75.8&82.7&88.2&63.7&43.3&61.8&45.6&49.9&62.4\\
    \hline
    \hline
    HiCOMEX   & 50.2&	\textbf{55.3}	&\textbf{58.2}&	\textbf{79.9}	&73.8&86.9&	\textbf{90.7}&	66.8	&\textbf{45.5}&	62.4	&46.1	&48.7&	\textbf{63.7} \\
    \hline
    \end{tabular}
    \end{table*}

  \begin{table*}[th]
   \caption[f1_disfa]{F1-score(\%)  in a comparative study of different state-of-the-art detection methods on the DISFA dataset.  
   To ensure fairness in this, we did not list some results found using BP4D pre-training, e.g. JAA~\cite{shao2020jaa}.}\label{tab:f1_disfa}
   \centering
   \begin{tabular}{c|c|c|c|c|c|c|c|c|c}
   \hline
   \hline
   AU & 1 & 	2	 & 4	 & 6	& 9	& 12	&  25	& 26& Avg\\
   \hline
   \hline
 LSVM&10.8&10.0&21.8&15.7&11.5&70.4&12.0&22.1&21.8\\
APL&11.4&12.0&31&12.4&10.1&65.9&21.4&26.9&23.8\\
DRML&17.3&17.7&37.4&29.0&10.7&37.7&38.5&20.1&26.7\\
EAC-Net&41.5&26.4&66.4&50.7&\textbf{80.5}&\textbf{89.3}&88.9&15.6&48.5\\
DSIN&42.4&39.0&68.4&28.6&46.8&70.8&90.4&42.2&53.6\\
CMS&40.2&44.3&53.2&\textbf{57.1}&50.3&73.5&81.1&59.7&57.4\\
LP-Net&29.9&24.7&72.7&46.8&49.6&72.9&93.8&65.0&56.9\\
ARL&43.9&42.1&63.6&41.8&40.0&76.2&\textbf{95.2}&\textbf{66.8}&58.7\\
   \hline
   \hline
HiCOMEX  &   \textbf{61.1}  & \textbf{58.2}&   \textbf{71.0}&   32.5   &54.2&   70.8&   93.8 &  52.4  &\textbf{61.8} \\
   \hline
   \end{tabular}
   \end{table*}

   \begin{table*}[th]
    \caption[f1_bp4d_ab]{F1-score(\%)  in an ablation study of HiCOMEX on the BP4D dataset.}\label{tab:f1_bp4d_ab}
    \centering
    \resizebox{\textwidth}{20mm}{\begin{tabular}{c|c|c|c|c|c|c|c|c|c|c|c|c|c}
    \hline
    \hline
    AU & 1 & 	2	 & 4	 & 6	& 7	& 10	& 12	& 14& 	15	& 17	& 23	& 24& Avg\\
    \hline
    \hline
    HiCOMEX w/o  intensity &  & 	 & 	 &	& 	&	& 	& & 	& 	& 	& & \\ distribution learning and  &50.6	&50.8	&56.5	&79.9	&\textbf{74.7}	&\textbf{88.2}	&90.3	&\textbf{68.5}	&40.2	&\textbf{63.5}	&43.7	&44.2 & 62.6 \\ COMEX learning  &  & 	 & 	 &	& 	&	& 	& & 	& 	& 	& &\\
    \hline
    HiCOMEX w/ only &  & 	 & 	 &	& 	&	& 	& & 	& 	& 	& & \\COMEX learning (self-attention) &51.1&	51.7	&59.0&	78.0&	73.4	&84.9	&89.9	&64.5	&\textbf{50.8}	&61.9	&47.2	&\textbf{48.8}  &63.4  \\
    \hline
 HiCOMEX w/ only &  & 	 & 	 &	& 	&	& 	& & 	& 	& 	& & \\COMEX learning (Hopfield) & 52.1&	53.8	&\textbf{60.7}&	76.3	&73.3&	87.1&	90.1&	64.0	&45.1&	63.3	&\textbf{48.5}	&45.6&	63.3 \\
    \hline
    HiCOMEX w/ only intensity&  & 	 & 	 &	& 	&	& 	& & 	& 	& 	& & \\  distribution  learning (LSTM) &\textbf{53.1}&	52.9	&59.8&	79.0&	73.2&	88.0	&90.6	&61.3&	46.2&	63.4	&48.1&	46.9&	63.5 \\
    \hline
    HiCOMEX   & 50.2&\textbf{55.3}	&58.2&	\textbf{79.9}	&73.8&	86.9&	\textbf{90.7}&	66.8	&45.5&	62.4	&46.1	&48.7&	\textbf{63.7} \\
    \hline
    \end{tabular}}
  \end{table*}

   \begin{table*}[th]
   \caption[f1_disfa_ab]{F1-score(\%)  in an ablation study of HiCOMEX on the DISFA dataset.  }\label{tab:f1_disfa_ab}
    \centering
   \begin{tabular}{c|c|c|c|c|c|c|c|c|c}
    \hline
    \hline
   AU & 1 & 	2	 & 4	 & 6	& 9	& 12	&  25	& 26& Avg\\
    \hline
    \hline
    HiCOMEX w/o  intensity &  & 	 & 	 &	& 	&	& 	& & \\ distribution learning and  &58.1	&52.2	&70.8	&20.6	&47.9	&70.4	&91.4	&\textbf{62.8}	&59.3
 \\ COMEX learning  &  & 	 & 	 &	& 	&	& 	& &\\
    \hline
    HiCOMEX w/ only &  & 	 & 	 &	& 	&	& 	& &  \\COMEX learning (self-attention)  & 60.1 &	\textbf{59.2}	 &\textbf{71.3}	 &26.5	 &\textbf{55.3}	 &71.5 &	93.8 &	53.1	 &61.4 \\
    \hline
 HiCOMEX w/ only &  & 	 & 	 &	& 	&	& 	& &  \\COMEX learning (Hopfield) & 61.0&	57.2&	70.6	&32.4	&51.4&	70.7&	90.7&	51.1 &	60.6 \\
    \hline
    HiCOMEX w/ only intensity&  & 	 & 	 &	& 	&	& 	& &\\  distribution  learning (LSTM)    & 59.3	&59.0	&69.5&	\textbf{35.4}	&46.2&	\textbf{72.2}	&\textbf{94.1}&	48.3	 &60.5  \\
    \hline
    HiCOMEX   &   \textbf{61.1}  &58.2&   71.0&   32.5   &54.2&   70.8&   93.8 &  52.4  &\textbf{61.8} \\
    \hline
    \end{tabular}
  \end{table*}

  \section{Experiments}
  \label{sec:experiments}
  
  \subsection{Dataset and evaluation metrics}
  \label{ssec:dataset}
  
Our HiCOMEX is evaluated on two widely used  datasets for assessing 
AU detection efficacy, i BP4D~\cite{zhang2014bp4d} and DISFA~\cite{mavadati2013disfa}, in which both AU and landmark labels are provided.

  BP4D contains 41 subjects, including 23 women and 18 men.
  Each involves eight sessions. There are 328
  videos including about 140,000 frames with AU labels.
  49 landmarks are provided for each frame. Akin to the setting of Zhao et 
  al.~\cite{zhao2016deep}, Li et al.~\cite{li2017eac}, and
   Shao et al.~\cite{shao2020jaa}, 12 AUs (1, 2, 4, 6, 7, 10, 12
  14, 15, 17, 23, and 24) are assessed  using subject exclusive
  three-fold cross-validation, where two folds are used
  for training and the remaining one is used for testing.
  
  DISFA contains 27 video recordings from 12 women
  and 15 men, each has 4845 frames. Each frame
  is labelled with its AU intensity in the range from 0 to 5, and 66 landmarks are provided. 
  According to
  the setting of Zhao et al.~\cite{zhao2016deep}, Li et al.~\cite{li2017eac}, and
  Shao et al.~\cite{shao2020jaa}, an AU intensity
   equal to or greater than 2 is considered as an occurrence
  (others are treated as non-occurrences).
  Eight AUs (1, 2, 4, 6, 9, 12, 25, and 26) are evaluated.

It should be noted  that in our experiments, we do not use any additional data or 
  pre-trained models. That is to say, we only use BP4D and DISFA's own data for 
  training and test.
  
  Following the previous methods of
  Zhao et al.~\cite{zhao2016deep}, Li et al.~\cite{li2017eac}, and
  Shao et al.~\cite{shao2020jaa}, 
  F1-score (\%) are reported for comparison.

  \subsection{Results and Discussions}
  \label{ssec:results}

    In this experiment, HiCOMEX is compared with several classical and state-of-the-art approaches, such as 
  LSVM~\cite{fan2008liblinear}, JPML~\cite{zhao2016joint}, DRML~\cite{zhao2016deep}, CPM~\cite{zeng2015confidence}
   EAC-Net~\cite{li2017eac}, DSIN~\cite{corneanu2018deep}, CMS~\cite{sankaran2019representation}, 
   LP-Net~\cite{niu2019local}, 
    ARL~\cite{shao2019facial}, SCC~\cite{fan2020facial}, and J$\hat{\text{A}}$A-Net ~\cite{shao2020jaa}.

  Tables~\ref{tab:f1_bp4d} and~\ref{tab:f1_disfa}  lists the results obtained by our methods and almost all of the results from methods developed in the past three years. 
 Compared 
    with these baselines, HiCOMEX obtained an absolute advantage, once again surpassing the performance
    of state-of-the-art systems. 
 HiCOMEX  is (in absolute terms) 1.3\% and 3.1\% better with an  F1-score  that beats the previous state-of-the-art method, respectively on BP4D and DISFA datasets.
    Especially on AU2, AU10, AU12, and AU14, the performances are much better than the previous best method. 
    These four AUs have many mutually  interactive relationships with other AUs~\cite{li2019semantic}. HiCOMEX uses this interactive 
    relationship to strengthen the discriminative ability on  AU2, AU10, and AU14 feature maps; our method still has absolute advantages 
    compared with other SOTA graph-based methods, such as SCC~\cite{fan2020facial} in Table~\ref{tab:f1_bp4d}.

  \subsection{Ablation study}
  \label{ssec:ablation}

  In order to prove the effectiveness of the key modules in HiCOMEX, 
   we undertook an  ablation study in five settings: HiCOMEX without either an intensity distribution  or COMEX learning, 
   HiCOMEX with only intensity distribution  learning (LSTM),  HiCOMEX with only COMEX learning (self-attention), 
   HiCOMEX with only COMEX learning (Hopfield), and the complete HiCOMEX.
  Tables~\ref{tab:f1_bp4d_ab}  and~\ref{tab:f1_disfa_ab} list the results. Use of the BP4D  
  intensity distribution  and COMEX learning will boost the performance from 62.6\% to 63.4\% and 63.5\% in  terms of F1-score respectively
  compared with the HiCOMEX version without both an intensity distribution  and COMEX learning.
  The full HiCOMEX pipeline performs best compared with all other versions.  
  That means the  local AU intensity distribution  and COMEX relationship-learning modules can boost the performance.
  The effect of all local AU relationship-learning methods is to increase  the F1-score of the  pipeline shown in 
  Fig.~\ref{detection_pipeline} 
  without all relationship-learning modules  by 1.1\%  on BP4D. The situation is similar to DISFA, the difference is that self-attention has the best effect.

  \section{CONCLUSION}
  \label{sec:conclusion}
  
We investigated the effectiveness of local AU feature map relationship modelling for 
  AU detection. We propose the use of HiCOMEX to undertake
  AU recognition. Benefits from the strength of AU intensity distribution  and COMEX relation learning, 
  the best performance of  HiCOMEX achieves the new state-of-the-art of 63.7\% and  61.8\%  F1-score on 
  the public BP4D and DISFA
  datasets respectively: these scores are achieved  without any use of  any  data or  pre-trained models.

\bibliographystyle{splncs03}
\bibliography{au_arxiv}

\end{document}